\documentclass{ifacconf}

\usepackage{graphicx}      
\usepackage{natbib}        
\usepackage{amsfonts}
\usepackage{amsmath}
\usepackage{xcolor}
\usepackage{subfig}
\usepackage{comment}
\usepackage{optidef}
\usepackage{caption}
\usepackage{enumitem}
\usepackage{array}
\usepackage{multirow}
\usepackage{tikz}
\usetikzlibrary{positioning,shapes}
\newcommand{\PreserveBackslash}[1]{\let\temp=\\#1\let\\=\temp}

\newcommand{\Compactcdots}{\mathinner{\cdotp\mkern-2mu\cdotp\mkern-2mu\cdotp}}
\DeclareMathSymbol{\shortminus}{\mathbin}{AMSa}{"39}

\DeclareMathOperator{\covar}{Cov}
\DeclareMathOperator{\trace}{Tr}

\newcommand{\eye}{%
	\text{\usefont{U}{bbold}{m}{n}1}%
}
\newcommand{\eyed}[1]{%
	\eye_{#1}%
}
\newcommand{\dpartial}[2]{\frac{\partial #1}{\partial #2}}

\newcommand{\bbR}{\mathbb{R}}
\newcommand{\humanUncEllip}{W^\mathrm{h}}
\newcommand{\statecov}{\Sigma}
\newcommand{\numStd}{\gamma}
\newcommand{\safeR}{\Delta_{\text{safe}}}
\newcommand{\DistRH}{\Delta_{\text{rh}}}

\newcommand{\curState}{\bar{x}_{0}}
\newcommand{\discInterval}{\delta_t}

\newcommand{\yunfan}[1]{\textcolor{black}{#1}}

\begin{document}
\begin{frontmatter}

\title{Stochastic Model Predictive Control with Optimal Linear Feedback for Mobile Robots in Dynamic Environments}

\thanks[footnoteinfo]{The research that led to this paper was funded by Robert Bosch GmbH. This work was also supported by
	DFG via Research Unit FOR 2401 and project 424107692 on Robust MPC and by the EU via ELO-X 953348.}

\author[First,Second]{Yunfan Gao},
\author[Second]{Florian Messerer},
\author[First]{Niels van Duijkeren},
\author[Second]{Moritz Diehl}

\address[First]{Robert Bosch GmbH, Corporate Research, Stuttgart, Germany
	(e-mail: \{yunfan.gao, niels.vanduijkeren\}@de.bosch.com).}
\address[Second]{Department of Microsystems Engineering (IMTEK), University of Freiburg, Germany
	(e-mail: \{florian.messerer, moritz.diehl\}@imtek.uni-freiburg.de)}

\begin{abstract}                
	Robot navigation around humans can be a challenging problem since human movements are hard to predict.
	Stochastic model predictive control (MPC) can account for such uncertainties and approximately bound the probability of a collision to take place.
	In this paper, to counteract the rapidly growing human motion uncertainty over time,
	  we incorporate state feedback in the stochastic MPC.
	This allows the robot to more closely track reference trajectories.
	To this end the feedback policy is left as a degree of freedom in the optimal control problem.
	The stochastic MPC with feedback is validated in simulation experiments and is compared against nominal MPC and stochastic MPC without feedback.
    The added computation time can be limited by reducing the number of additional variables for the feedback law with a small compromise in control performance.
\end{abstract}

\vspace{-7pt}
\begin{keyword}
robotics, stochastic model predictive control, motion planning
\end{keyword}

\end{frontmatter}

\section{Introduction}
\vspace{-6pt}

Increasingly many autonomous mobile robots (AMRs) operate alongside humans in industry.
A fundamental prerequisite for AMRs to work in a shared space with humans is that they do not collide with humans~\citep{Brock2002}.
In order to coordinate robot motions effectively,
   it is essential to consider predictions of human trajectories, which are generally uncertain.
Uncertainty-aware planning often includes probabilistic constraints for collision avoidance~\citep{DuToit2012, Carvalho2014, Zhu2019}.
A real-time capable approach for dynamic obstacle avoidance using stochastic model predictive control (MPC) is presented in~\citep{Castillo-Lopez2020}.

Human motion uncertainty typically grows rapidly over time.
Large uncertainties cause the robot to keep an overly big distance to the humans,
    potentially deviating from intended reference trajectories.
Incorporating a feedback law in the optimal control problem (OCP) formulation can mitigate conservativeness~\citep{Scokaert1998}.
The scheme can be crudely approximated by keeping the covariance matrices constant after some time steps in the future~\citep{Yan2005} or bounding the matrices by pre-defined values~\citep{Schwarting2018}.
These approaches, however, overlook the resulting uncertainty of robot state and the discrepancy in the robot feedback capacities at different robot states.

In robust and stochastic MPC, several research works explicitly model the feedback policies parameterized by, e.g., linear feedback laws~\citep{Nagy2004}.
The associated feedback gain matrices can either be precomputed~\citep{Mayne2011a} or optimized online~\citep{Messerer2021}.
For motion planning of nonhonolomic robot systems,
    e.g., differential-drive or tricycle robots,
    precomputed constant feedback gains are unlikely to be effective for all combinations of robot states and human positions.
Therefore, treating the gain matrices as decision variables is especially beneficial.
Meanwhile, it results in optimization problems that are in general challenging to solve~\citep{Goulart2006}.
Incorporating feedback on human motion uncertainty in motion planning with the feedback gain matrices being optimized has not yet been extensively investigated in the existing literature.

In this paper, we present a stochastic MPC formulation for robot motion planning;
    focusing on Gaussian-distributed human motion uncertainty
    and incorporating feedback on the uncertainty.
By probabilistically avoiding collisions with people in the MPC prediction horizon and enforcing the robot to almost stand still after that,
    we aim for a concept called \textit{passive safety}~\citep{Macek2008}.
While the feedback can reduce the uncertainty of the human-robot distance,
    it increases the uncertainty of the robot state.
Therefore, it is desired that feedback is weaker when the human is further away.
We accomplish this through optimizing a time-varying linear feedback law in the OCP,
    coupled with carefully designed cost terms and constraints in the OCP formulation.

We discuss and compare the usage of two feedback policies and a baseline without feedback.
One feedback policy considers both the robot and human state uncertainty while the other considers only the human.
The performance and the solution time are presented,
    highlighting the possibility to trade off computation time with control performance.

\vspace{-2pt}
\subsubsection{Notation}
The nominal system state is denoted by $x$ and the disturbed system state is denoted by $\tilde{x}$.
Let $\bar{x}$ be the currently measured system state.
An all-ones vector of the size $n$ is written as $\eyed{n}$.
The notation of a zero matrix of the size $m \times n$ is $[0]_{m \times n}$.
A set of natural numbers for the interval $\left[a, b \right]$, possibly starting from~0, is denoted by $\mathbb{N}_{\left[a, b \right]}$.
The quadratic norm weighted by a symmetric positive definite matrix $Q$ is denoted by $\|x\|^2_Q:=x^\top Qx$.
A multi-variable Gaussian distribution with mean $\mu$ and variance $\Sigma$ is represented by $\mathcal{N}\left( \mu, \Sigma\right)$.

\vspace{-3pt}
\section{Model Predictive Control Problem}
\vspace{-6pt}
\label{sec:problem-statement}
\subsubsection*{Robot Model}
Consider a kinematic model of a differential-drive robot with center position $\left( p_x^{\text{r}}, p_y^{\text{r}}\right)$ and heading~$\theta$~\citep{Siegwart2011}.
The forward and angular velocities are denoted by~$v^{\text{r}}$ and $\omega$, respectively:
\begin{equation}
	x^{\text{r}} := \left[ \begin{array}{ccccc}
		p_x^{\text{r}} & p_y^{\text{r}} & \theta & v^{\text{r}} & \omega
	\end{array}\right]^\top \in \bbR^5.
	\label{eq:robot_state}
\end{equation}
The control input $u^{\text{r}}$ consists of forward acceleration~$a^{\text{r}}$ and angular acceleration $\alpha$, i.e.,
$u^{\text{r}} := \left[ \begin{array}{cc}
	a^{\text{r}} & \alpha
\end{array} \right]^{\top}\in \bbR^2. $
The continuous-time equations of motion are
\begin{equation}
	\dot{x}^{\text{r}} = \left[ \begin{array}{ccccc}
		v^{\text{r}}\cos(\theta) & v^{\text{r}}\sin(\theta) & \omega & a^{\text{r}} & \alpha
	\end{array}\right]^\top .
	\label{eq:diff-drive-cont-time-dynamics}
\end{equation}
Consider a zero-order-hold control signal with discretization intervals of length $\delta_t$.
The system is discretized by numerical integration (explicit Runge-Kutta integrator of order four):
\begin{equation}
	x_{k+1}^{\text{r}} = \psi_k^{\text{r}}(x_k^{\text{r}}, u_k^{\text{r}})\colon \bbR^{5}\times\mathbb{R}^{2}\to \mathbb{R}^{5}.
\end{equation}
The robot is subject to affine constraints on $v^{\text{r}}, \omega, a^{\text{r}}$, and $\alpha$ due to actuator limits.
The nonlinear collision avoidance constraints will be detailed in a later paragraph.

\vspace{-3pt}
\subsubsection*{Human Model}
\yunfan{The human model, as we will see in Section~\ref{sec:cov-prop},
    plays a role in the uncertainty propagation of the system state when the control policy incorporates feedback on human motion uncertainty.}
The state of the human is its center position and its input is the velocity:
\begin{equation}
	x^{\text{h}} := \left[ \begin{array}{cc}
		p_x^{\text{h}} & p_y^{\text{h}}
	\end{array}\right]^\top,
    \ u^{\text{h}} := \left[ \begin{array}{cc}
    	v_x^{\text{h}} & v_y^{\text{h}}
    \end{array}\right]^\top,
    \ \text{and} \   \dot{x}^{\text{h}} = u^{\text{h}}.
    \label{eq:human_state}
\end{equation}
The input of the human motion $\tilde{u}^{\text{h}}(t)$ is also modeled as a zero-order hold signal with the same step length $\delta_t$.
The input $\tilde{u}^{\text{h}}_k$ is assumed to be Gaussian-distributed:
\begin{equation*}
	\tilde{u}^{\text{h}}_k ~\sim \mathcal{N}\left( {u}^{\text{h}}_k , \humanUncEllip_k \right), \ \text{for} \ k \in\mathbb{N}_{\left[0, N \shortminus 1 \right]},
\end{equation*}
where mean ${u}^{\text{h}}_k$ and covariance $\humanUncEllip_k \in \bbR^{2\times 2}$ are time-varying.
The discrete-time system dynamics are
\begin{equation}
\tilde{x}^{\text{h}}_{k+1} = A^{\text{h}}\tilde{x}^{\text{h}}_k + B^{\text{h}}\tilde{u}^{\text{h}}_k, \ \text{for} \ k \in\mathbb{N}_{\left[0, N \shortminus 1 \right]}.
\end{equation}
We assume that the current human state is measured without uncertainty.
The human does not react to the robot movement.
The human velocity prediction and covariance estimates across any two subsequent predictions are consistent.

\vspace{-3pt}
\subsubsection*{Collision Avoidance}

Denote the distance between the robot position and the human position as
\begin{equation*}
	\DistRH\left( x^{\text{r}}, x^{\text{h}}\right)  := \left\| \left[
	\begin{array}{cc}
		p_x^{\text{r}} & p_y^{\text{r}}
	\end{array} \right]^\top
	- \left[
	\begin{array}{cc}
		p_x^{\text{h}} & p_y^{\text{h}}
	\end{array} \right]^\top  \right\|_2.
\end{equation*}
The distance shall not be less than a minimum collision-free distance $\safeR$, i.e.,
\begin{equation}
	\DistRH\left( x^{\text{r}}, x^{\text{h}}\right) - \safeR \geq 0.
	\label{eq:coll-avoid-def}
\end{equation}

\vspace{-3pt}
\subsubsection*{Terminal Constraints}
We impose a constraint requiring the robot to be nearly stationary at the terminal state:
\begin{equation}
	0 \leq v^{\text{r}}_N \leq \epsilon_v.
	\label{eq:terminal-constr-velocity}
\end{equation}
The upper bound is a small positive constant $\epsilon_v$.
The control policy beyond the terminal state is to remain standing still.
The terminal constraints together with the stabilizing control policy ensure that the robot is passively safe, i.e., the robot does not collide while moving.
Here, the robot is allowed to have a non-zero terminal angular velocity due to its circular shape.

\vspace{-3pt}
\subsubsection*{Reference Trajectory Tracking}
We assume that state and input reference trajectories are given:
\begin{equation}
	\left( x^{\mathrm{r,ref}}_{0}, \dots, x^{\mathrm{r,ref}}_{N}  \right) \ \text{and} \
	\left( u^{\mathrm{r,ref}}_{0}, \dots, u^{\mathrm{r,ref}}_{N-1} \right)
	\label{eq:reference_trajectory-def}.
\end{equation}
The stage and terminal costs of the OCP are weighted squared tracking errors:
\begin{subequations}
	\begin{align}
		l_k(x_k^{\mathrm{r}}, u_k^{\mathrm{r}})\coloneqq
		& \frac{1}{2}\|u_k\!-\!u^{\mathrm{r,ref}}_{k}\|_R^2 + \frac{1}{2}\|x_k\!-\!x^{\mathrm{r,ref}}_{k}\|_Q^2,
		\label{eq:tracking-stage-def} 		\\
		l_N(x_N^{\mathrm{r}})\coloneqq
		& \frac{1}{2}\|x_N^{\mathrm{r}}\!-\!x^{\mathrm{r,ref}}_{N}\|_{Q_{\mathrm{e}}}^2.
	\end{align}
    \label{eq:tracking-def}
\end{subequations}

\vspace{-6pt}
\prob{
Given predictions of a human motion trajectory and human motion uncertainty,
    the aim is to find a control policy incorporating \textit{feedback} on human motion uncertainty
such that the robot tracks a reference trajectory~\eqref{eq:reference_trajectory-def}
    while probabilistically satisfying the input-state constraints and collision avoidance constraints~\eqref{eq:coll-avoid-def}.

\section{Stochastic OCP with Optimized Linear State Feedback}
\vspace{-6pt}
The system state $x\in\bbR^{7}$ consists of the concatenation of the robot and the human states.
The control input in the OCP is the actual input $u^{\text{r}}$ of the robot.
The input of the human motion model is considered as a stochastic disturbance.
The control policy incorporates feedback on the state deviation $\left(  \tilde{x}_k - {x}_k \right) $.
The feedback scheme is parameterized as a time-varying linear feedback policy:
\begin{equation}
	\tilde{u}_k^{\text{r}}(\tilde{x}_k) := {u}_k^{\text{r}} + K_k\left( \tilde{x}_k - {x}_k \right),
	\label{eq:linear-state-feedback}
\end{equation}
with $K_k\in\bbR^{2 \times 7}$ for $k \in \mathbb{N}_{\left[1, N \shortminus 1\right]}$.
For $k=0$, we do not consider a feedback gain because the current human state is assumed to be known exactly.

\vspace{-3pt}
\subsection{Covariance Propagation}
\vspace{-7pt}
\label{sec:cov-prop}
The state covariance matrices $\statecov_{k}$
consist of the robot state covariance $\statecov_{k}^{\mathrm{r}}\in\bbR^{5\times 5}$ and the human state covariance $\statecov_{k}^{\mathrm{h}}\in\bbR^{2\times 2}$ as diagonal blocks.
The off-diagonal blocks $\statecov_{k}^{\mathrm{rh}}\in\bbR^{5\times 2}$ represent the covariance between the robot and human states:
\begin{equation*}
	\statecov_{k} := \left[ \begin{array}{cc}
		\statecov_{k}^{\mathrm{r}} & \statecov_{k}^{\mathrm{rh}} \\
		\left\lbrace \statecov_{k}^{\mathrm{rh}}\right\rbrace ^{\!\top} & \statecov_{k}^{\mathrm{h}}
	\end{array}
	\right].
\end{equation*}
The covariance propagation can be approximated based on the linearization of the discretized system dynamics at the nominal trajectory.
Let
\begin{subequations}
	\begin{align}
		A_k^{\mathrm{r}}    \! &   := \! \dpartial{\psi_k^{\mathrm{r}}}{x_k^{\mathrm{r}}}(x_k^{\mathrm{r}}, u_k^{\mathrm{r}}),
		\quad B_k^{\mathrm{r}}  \! := \! \dpartial{\psi_k^{\mathrm{r}}}{u_k^{\mathrm{r}}}(x_k^{\mathrm{r}}, u_k^{\mathrm{r}}), \\
		\check{A}_k  \! & :=  \! \left[ \begin{array}{cc}
			A_k^{\mathrm{r}} & {[0]}_{5 \times 2} \\ {[0]}_{2 \times 5} & A^{\mathrm{h}}
		\end{array} \right]  +
		\left[ \begin{array}{c}
			B_k^{\mathrm{r}} \\ {[0]}_{2 \times 2}
		\end{array} \right] K_k,
	\end{align}
\end{subequations}
for $k \in \mathbb{N}_{\left[0, N \shortminus 1\right]}$.
The covariance propagation is given by
\begin{equation}
	\begin{split}
		\statecov_{k+1} =&  \, \Phi_k(x_k^{\mathrm{r}}, u_k^{\mathrm{r}}, \statecov_k, K_k) \\
		:=& \, \check{A}_k \statecov_k \check{A}_k^\top +
		\left[ \begin{array}{c}
			{[0]}_{5 \times 2}  \\ B^{\mathrm{h}}
		\end{array} \right] W_k^{\mathrm{h}} \left[ \begin{array}{c}
			{[0]}_{5 \times 2}  \\ B^{\mathrm{h}}
		\end{array} \right]^{\!\top},
	\end{split}
	\label{eq:uncertainty_prop_wFdbk}
\end{equation}
for $k \in \mathbb{N}_{\left[0, N \shortminus 1\right]}$.
Note that the covariance propagation of the human state uncertainty, i.e.,
\begin{equation*}
	\statecov_{k+1}^{\mathrm{h}} = A^{\mathrm{h}}\statecov_{k}^{\mathrm{h}}\left\lbrace A^{\mathrm{h}}\right\rbrace ^\top  + B^{\mathrm{h}} W_k^{\mathrm{h}} \left\lbrace B^{\mathrm{h}}\right\rbrace ^\top,
\end{equation*}
depends on neither the robot nominal trajectory nor the feedback gain matrices.
Therefore, the uncertain human trajectory can be precomputed and passed to the OCP as parameters.
The robot state uncertainty $\statecov_{k}^{\mathrm{r}}$ and the robot-human covariance $\statecov_{k}^{\mathrm{rh}}$ are propagated inside the OCP:
\begin{subequations}
	\begin{align}
		\statecov_{k+1}^{\mathrm{r}} &= \, \Phi_k^{\mathrm{r}}(x_k^{\mathrm{r}}, u_k^{\mathrm{r}}, \statecov_k, K_k)\\
		&:= \left[ \begin{array}{cc}
			{I}_{5} & {[0]}_{5 \times 2}
		\end{array} \right]
		\Phi_k(x_k^{\mathrm{r}}, u_k^{\mathrm{r}}, \statecov_k, K_k)
		\left[ \begin{array}{cc}
			{I}_{5} & {[0]}_{5 \times 2}
		\end{array} \right]^\top \!, \nonumber \\
		\statecov_{k+1}^{\mathrm{rh}} &= \, \Phi_k^{\mathrm{rh}}(x_k^{\mathrm{r}}, u_k^{\mathrm{r}}, \statecov_k, K_k) \\
		&:= \left[ \begin{array}{cc}
			{I}_{5} & {[0]}_{5 \times 2}
		\end{array} \right]
		\Phi_k(x_k^{\mathrm{r}}, u_k^{\mathrm{r}}, \statecov_k, K_k)
		\left[ \begin{array}{cc}
			{[0]}_{2 \times 5} & {I}_{2}
		\end{array} \right]^\top\!, \nonumber
	\end{align}
\end{subequations}
for $k \in \mathbb{N}_{\left[0, N \shortminus 1\right]}$.

\vspace{-3pt}
\subsection{Expected Stage Cost and Terminal Cost}
\vspace{-8pt}
\label{sec:expected-cost-due-to-uncertainty}
The state and input uncertainties due to feedback policies affect the expected cost we aim to minimize.
Considering the Gaussian distribution of the state and the linear state feedback policy~\eqref{eq:linear-state-feedback},
we have
\begin{equation}
	\covar( \tilde{x}_k^{\mathrm{r}} ) = \statecov_{k}^{\mathrm{r}} \ \text{and} \ \covar( \tilde{u}_k^{\mathrm{r}} )  = K_k \statecov_k K_k^\top.
\end{equation}
Given the trajectory-tracking cost defined in~\eqref{eq:tracking-def},
the expected value of the stage cost is
\begin{equation}
	\begin{split}
		\mathbb{E}\left\lbrace l_k(\tilde{x}_k^{\mathrm{r}}, \tilde{u}_k^{\mathrm{r}})  \right\rbrace
		\!&=
			\trace \left\lbrace \frac{1}{2} Q
			\statecov_{k}^{\mathrm{r}} \right\rbrace
			\!+\! \trace \left\lbrace \frac{1}{2} R
			K_k \statecov_k K_k^\top
			\right\rbrace \\
			&\quad +  l_k({x}_k^{\mathrm{r}}, {u}_k^{\mathrm{r}})
			=: \tilde{l}_k\left(x_k^{\mathrm{r}}, u_k^{\mathrm{r}}, \statecov_{k}, K_k\right). \nonumber
	\end{split}
	\label{eq:expeced-cost-def}
\end{equation}
The expected value of the terminal cost $\tilde{l}_N\left(x_N^{\mathrm{r}},\statecov_{N}\right)$ can be defined similarly.
~\yunfan{We refer to~\citep{Messerer2021} for the derivation}.
\vspace{-3pt}
\subsection{Penalized Chance Constraints}
\vspace{-8pt}
Consider the stage constraints on the system state and input:
$ h_k(\tilde{x}_k, \tilde{u}_k)\leq 0: \bbR^{7} \!\times \! \bbR^{2} \!\to \!\bbR^{n_{h,k}}$ .
Based on the linearization of the constraints on the nominal trajectory $\left(x_k, u_k \right)$,
the variance of one constraint component $h_{k,i}$ can be approximated by
\begin{equation*}
	\begin{split}
		\beta_{k,i} = &\, H_{k,i}(x_k, u_k, \statecov_{k}, K_k) \\
		:= &\, \nabla h_{k,i}(x_k, u_k)^{\! \top}
		\left[\begin{array}{c}
			{I}_{7}\!\\ K_k
		\end{array} \right] \! \statecov_k
		{\left[\begin{array}{c}
				{I}_{7}\\ K_k
			\end{array} \right]}^{\!\top} \nabla h_{k,i}(x_k, u_k).
	\end{split}
\end{equation*}
The constraint variance $\beta_N$ for terminal constraints can be defined similarly.
Due to the unbounded support of the Gaussian distribution, constraints can only be satisfied in a probabilistic manner~\citep{Schwarm1999}:
\begin{equation}
	h_{k,i}(x_k, u_k) + \numStd \sqrt{\beta_{k,i}} \leq 0,
	\label{eq:chance-constr-def}
\end{equation}
where $\numStd$ is the number of standard deviations that the nominal trajectory should keep from the nominal bounds.
Large disturbances can occasionally lead to infeasibility of state-dependent constraints of the OCP.
Soft constraints are therefore recommended in practice:
\begin{equation}
	\left\lbrace \begin{array}{ll}
		s_{k,i} & \geq h_{k,i}(x_k, u_k) + \numStd \sqrt{\beta_{k,i}}, \\
		s_{k,i} & \geq 0,
	\end{array}
	\right.
	\label{eq:chance-constr-soft-def}
\end{equation}
where $s_{k,i}$ are slack variables.
Weighted $l_1$ penalties $\tau_{k,i}s_{k,i}$ are added to the objective function.
If the weights $\tau_{k,i}$ are sufficiently large and the problem is feasible,
the solution of the softened problem corresponds to a solution of the problem with hard constraints~\citep{Han1979}.
In the infeasible case, the soft-constrained formulation will return a solution that minimizes constraint violation.

\rem{
	Incorporating feedback on human motion uncertainty can reduce the uncertainty of the relative position
	but meanwhile increases the uncertainty of the robot state.
	Therefore, the uncertainty of the relative position should only be reduced when necessary, i.e., when the collision avoidance constraints are active.
	The expected stage cost and terminal cost~\eqref{eq:expeced-cost-def} encourage smaller uncertainty in the robot state and robot input in general.
}

\vspace{-5pt}
\subsubsection*{Terminal Velocity Constraint}
It is desirable that the robot does in fact stand still at the end of the prediction horizon despite the uncertainty.
Instead of imposing the penalized chance constraint~\eqref{eq:chance-constr-soft-def} for the terminal forward velocity,
we impose the nominal velocity constraint~\eqref{eq:terminal-constr-velocity} and additionally restrict the variance of the forward velocity in the terminal state $\statecov_N^{v}$ to be less than a small $\epsilon_{\statecov}$ value:
\begin{equation}
	\statecov_N^{v} \leq \epsilon_{\statecov} \ll 1.
	\label{eq:constr-terminal-cov}
\end{equation}

\vspace{-5pt}
\subsection{OCP Formulation}
\vspace{-5pt}
The covariance matrices $\statecov_{k}$ may not remain positive (semi)definite throughout optimization iterations
when we treat the matrices $\statecov_{k}^{\mathrm{r}}$ and $\statecov_{k}^{\mathrm{rh}}$ as optimization variables in the OCP formulation.
This results in potentially negative values for the constraint uncertainties $\beta_k$ and issues when computing the square root~\eqref{eq:chance-constr-soft-def}.
The problem can be resolved by initializing $\beta_k$ with positive values
and imposing
\begin{equation*}
	\beta_{k} \geq \epsilon_{\beta} \eyed{n_{h,k}} \ \text{and} \ \beta_k \geq H_k(x_k, u_k, \statecov_k, K_k)
\end{equation*}
instead of $\beta_k = H_k(x_k, u_k, \statecov_k, K_k)$ with $\epsilon_{\beta}$ being a small positive value \citep{Messerer2023}.

Given the assumption that the human does not respond to the robot motion, the human nominal trajectory can be precomputed.
For compactness of notation, we define
\begin{equation*}
	{Y}_k^{\mathrm{r}} \!:=   \left( x_0^{\mathrm{r}},\! \Compactcdots, x_k^{\mathrm{r}}, u_0^{\mathrm{r}}, \!\Compactcdots, u_k^{\mathrm{r}},\! K_0,\! \Compactcdots,\! K_{k} \right),\ \text{for} \ k \in \mathbb{N}_{\left[0, N \shortminus 1 \right]}.
\end{equation*}
The stochastic OCP incorporating full-state feedback can be formulated as
\begin{mini!}|s|
	{\substack{{Y}_{N \!\shortminus\!1}^{\mathrm{r}}, \statecov_0^{\mathrm{r}}, \Compactcdots, \statecov_{N}^{\mathrm{r}}, \\
			\statecov_0^{\mathrm{rh}}, \Compactcdots, \statecov_{N}^{\mathrm{rh}}, x_N^{\mathrm{r}}, \\
			s_0, \Compactcdots s_N,
			\beta_0, \Compactcdots, \beta_N}}
	{\begin{array}{l}
			\sum_{k=0}^{N-1} \tilde{l}_k\left(x_k^{\mathrm{r}}, u_k^{\mathrm{r}}, \statecov_{k}, K_k\right)\\
			\quad  + \tilde{l}_N\left(x_N^{\mathrm{r}}, \statecov_{N}\right) + \sum_{k=0}^{N} \tau_{k}^{\top} s_k
	\end{array}}
	{\label{eq:SMPC-MS-CC}}{}
	\addConstraint{x_0^{\text{r}}}{=\curState^{\text{r}}, \, x_{k+1}^{\text{r}} = \psi_k^{\text{r}}(x_k^{\text{r}}, u_k^{\text{r}})}{}
	\addConstraint{\statecov_0^{\mathrm{r}}}{=[0]_{5 \times 5}, \, \statecov_0^{\mathrm{rh}}=[0]_{5 \times 2}}
	\addConstraint{K_0}{=[0]_{2 \times 7}}
	\addConstraint{\statecov_{k+1}^{\mathrm{r}}}{=\Phi_k^{\mathrm{r}}(x_k^{\mathrm{r}}, u_k^{\mathrm{r}}, \statecov_k, K_k) \label{eq:SMPC-cov-prop-r}}{}
	\addConstraint{\statecov_{k+1}^{\mathrm{rh}}}{=\Phi_k^{\mathrm{rh}}(x_k^{\mathrm{r}}, u_k^{\mathrm{r}}, \statecov_k, K_k) \label{eq:SMPC-cov-prop-rh}}{}
	\addConstraint{\beta_{k}}{\geq \epsilon_{\beta} \eyed{n_{h,k}}, \, \beta_{N} \geq \epsilon_{\beta} \eyed{n_{h,N}}}{}
	\addConstraint{\beta_k}{\geq H_k(x_k, u_k, \statecov_k, K_k) \label{eq:SMPC-stage-beta}}{}
	\addConstraint{\beta_N}{\geq H_N(x_N, \statecov_N) \label{eq:SMPC-terminal-beta}}{}
	\addConstraint{s_k}{\geq 0, \, s_N\geq 0}{}
	\addConstraint{s_k}{\geq h_k(x_k, u_k) + \gamma \sqrt{\beta_k} \label{eq:SMPC-MS-stage-ineq}}
	\addConstraint{s_N}{\geq h_N(x_N) +  \gamma \sqrt{\beta_N} \label{eq:SMPC-MS-terminal-ineq}}
	\addConstraint{\epsilon_v}{\geq v^{\text{r}}_N \geq 0,\,  \epsilon_{\statecov} \geq \statecov_N^{v} \label{eq:SMPC-MS-terminal-v}}
	\addConstraint{}{\quad \mathrm{with} \ k \in \mathbb{N}_{\left[0, N \shortminus 1 \right]} \nonumber},
\end{mini!}
where $\curState^{\text{r}}$ is the value of the current robot state.
The value of the current state uncertainty is set to zero because the current state is assumed to be known exactly.
The terminal velocity constraint is written separately in~\eqref{eq:SMPC-MS-terminal-v} and not included in~\eqref{eq:SMPC-MS-terminal-ineq}.

\vspace{-3pt}
\subsection{Feedback Policy Variations}
\vspace{-8pt}
The linear feedback policy defined in~\eqref{eq:linear-state-feedback} corresponds to incorporating feedback on both the robot state and the human state.
It will be referred to as \textit{full}-state feedback for the remainder of this paper.
Apart from the full-state feedback policy,
another option is to provide feedback solely on the human motion uncertainty, i.e.,
$ K_k^{\text{h}}\left( \tilde{x}_k^{\text{h}} - {x}_k^{\text{h}} \right)$,
with $K_k^{\text{h}} \in \bbR^{2 \times 2}$ being the feedback gain for the human state deviation.
The feedback law introduces uncertainty to the robot velocity,
due to which constraint~\eqref{eq:constr-terminal-cov} can be difficult to satisfy once the OCP formulation models feedback on the human state deviation.
Therefore, the robot needs to be able to reduce its own forward velocity uncertainty:
\begin{equation}
	\tilde{u}_k^{\text{r-PRTL}}(\tilde{x}_k)  :=  {u}_k^{\text{r}} \!+\! K_k^{\text{h}}\left( \tilde{x}_k^{\text{h}} - {x}_k^{\text{h}} \right) \!+\! K_k^{ \text{r}, v } \left(\tilde{v}_k^{\text{r}} - {v}_k^{\text{r}}\right),
	\label{eq:partial-fdbk}
\end{equation}
for $k \in \mathbb{N}_{\left[1, N \shortminus 1 \right]}$, where $K_k^{ \text{r}, v}$ is a scalar.
The feedback policy defined in~\eqref{eq:partial-fdbk} will be referred to as \textit{partial}-state feedback.

The capability of feeding back the robot forward-velocity uncertainty is particularly essential towards the end of the prediction horizon.
Therefore, it can be an implementation choice to allow non-zero $K_k^{ \text{r}, v }$ for only a subset of time instants.
The OCP formulation corresponding to the partial-state feedback includes only the non-zero feedback gain elements, i.e., $K_k^{\text{h}}$ and $K_k^{ \text{r}, v}$, in the optimization variables.

\rem{
Compared with the partial-state feedback policy,
treating the full feedback gain matrices as decision variables incurs an increase of optimization variables and the nonlinearity in the optimization problem.
On the other hand, it enables the robot to counteract its state uncertainty gained from the feedback on the human motion uncertainty.
Moreover, it can handle more general situations where the robot motion itself is subject to process noise.
}

Incorporating feedback contrasts with sticking to nominal input trajectories irrespective of future realizations of disturbance.
The former is referred to as \textit{closed-loop} stochastic OCP while the latter as \textit{open-loop} stochastic OCP for the remainder of this paper.
The open-loop stochastic OCP can also be regarded as a special case where $K_k = [0]_{2\times 7}$ for $k \in \mathbb{N}_{\left[1, N \shortminus 1 \right]}$.

\vspace{-3pt}
\section{Results}
\label{sec:results}
\vspace{-6pt}
\begin{figure*}[th]
	\vspace*{-10pt}
	\centering
	\captionsetup[subfloat]{captionskip=-2pt}
	\subfloat[Full-state feedback.]
	{\includegraphics[width=0.32\linewidth]
		{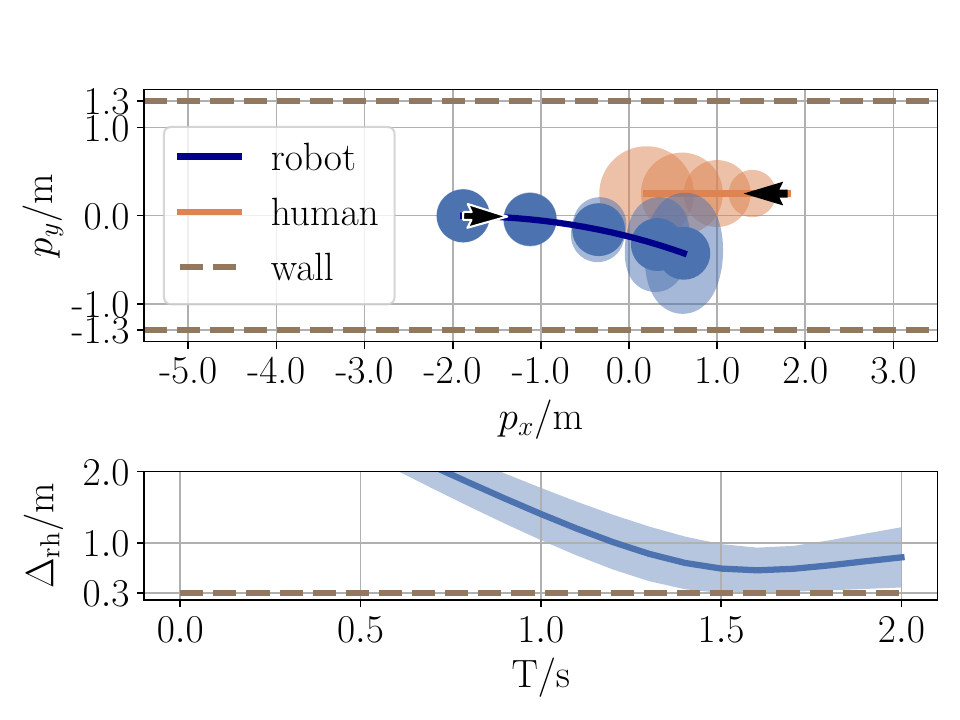}
		\label{fig:ocp_solution_OPT_corridor_space_whole_188}}
	\hfill
	\subfloat[Full-state feedback.]
	{\includegraphics[width=0.32\linewidth]
		{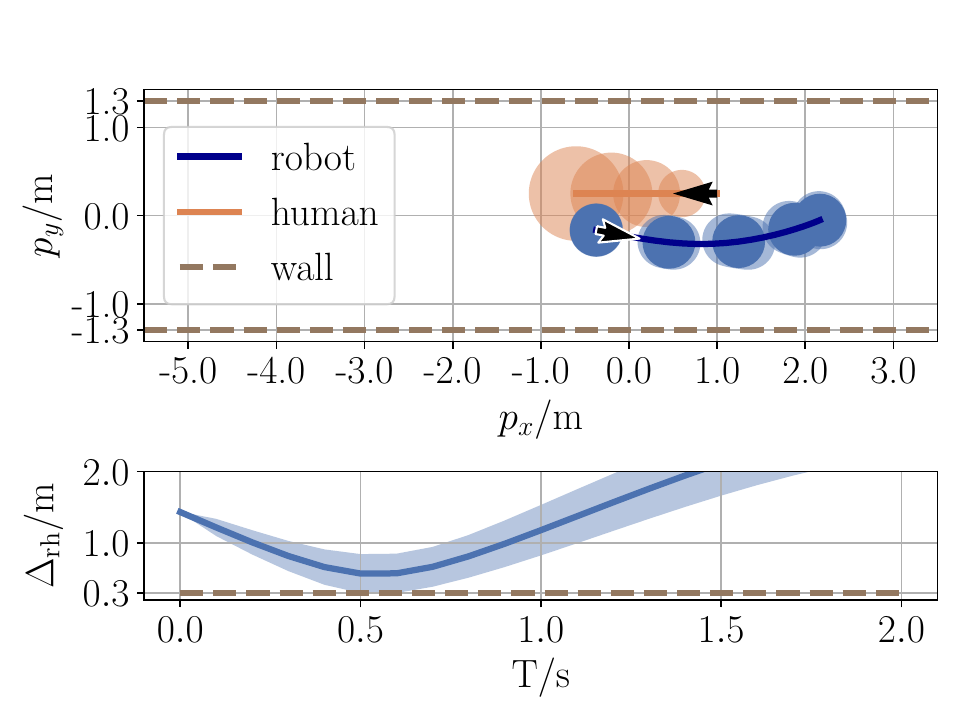}
		\label{fig:ocp_solution_OPT_corridor_space_whole_044}}
	\hfill
	\subfloat[Partial-state feedback.]
	{\includegraphics[width=0.32\linewidth]
		{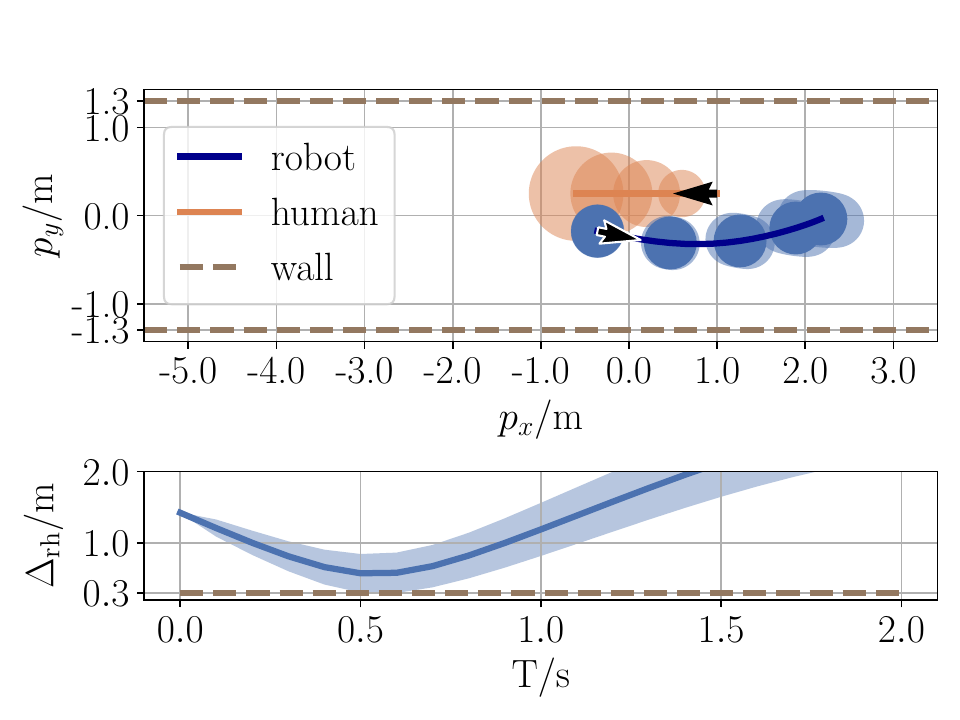}
		\label{fig:ocp_solution_OPT_corridor_space_partial_042}}\\
	\vspace{-5pt}
	\caption{Planned trajectories.
		The solid blue lines and solid orange lines in the upper plots represent the nominal trajectory of the robot and human respectively.
		The blue circles are of radius $\safeR$, and they correspond to time instant 0.0\,s, 0.5\,s, 1.0\,s, 1.5\,s, and 2.0\,s respectively.
		Each blue semi-transparent set depicts the Minkowski sum of a radius-$\safeR$ circle and the ellipsoid corresponding to the three standard deviations of the robot position uncertainty.
		The orange ellipsoids and the shadows in the distance plots represent uncertainty tubes (three standard deviations).
		}
	\label{fig:ocp_traj_in_space}
	\vspace{-15pt}
\end{figure*}

\begin{figure*}[th]
	\begin{minipage}[b]{0.59\linewidth}
		\centering
		\captionsetup[subfloat]{captionskip=-2pt}
		\subfloat[Robot and human meet in the later part of the prediction horizon (Fig.~\ref{fig:ocp_solution_OPT_corridor_space_whole_188}).]
		{\begin{tikzpicture}
				\node(a){\includegraphics[width=0.46\linewidth, trim={4pt 0 4pt 0},clip]
				{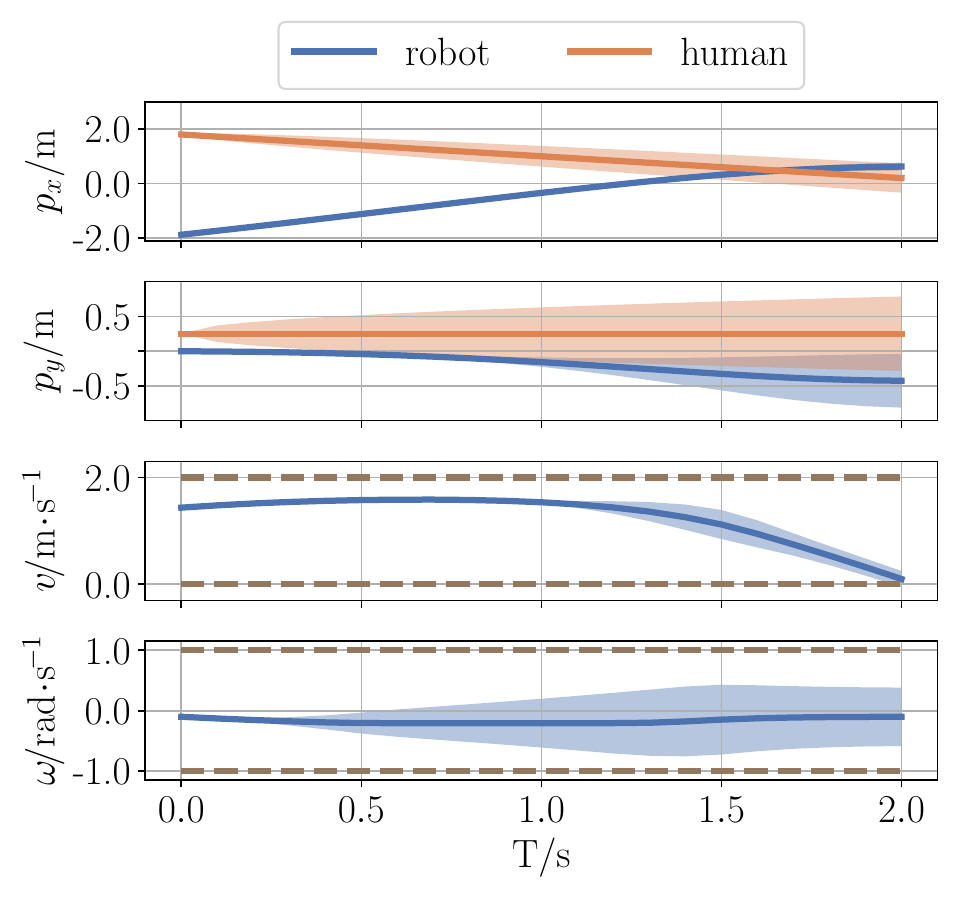}};
		\end{tikzpicture}
		\label{fig:ocp_solution_OPT_corridor_time_whole_188}
		}
	    \hfill
	    \subfloat[Robot and the human pass by within the prediction horizon. (Fig.~\ref{fig:ocp_solution_OPT_corridor_space_whole_044}).]
	    {\begin{tikzpicture}
	    	\node(a){\includegraphics[width=0.46\linewidth, trim={4pt 0 4pt 0},clip]{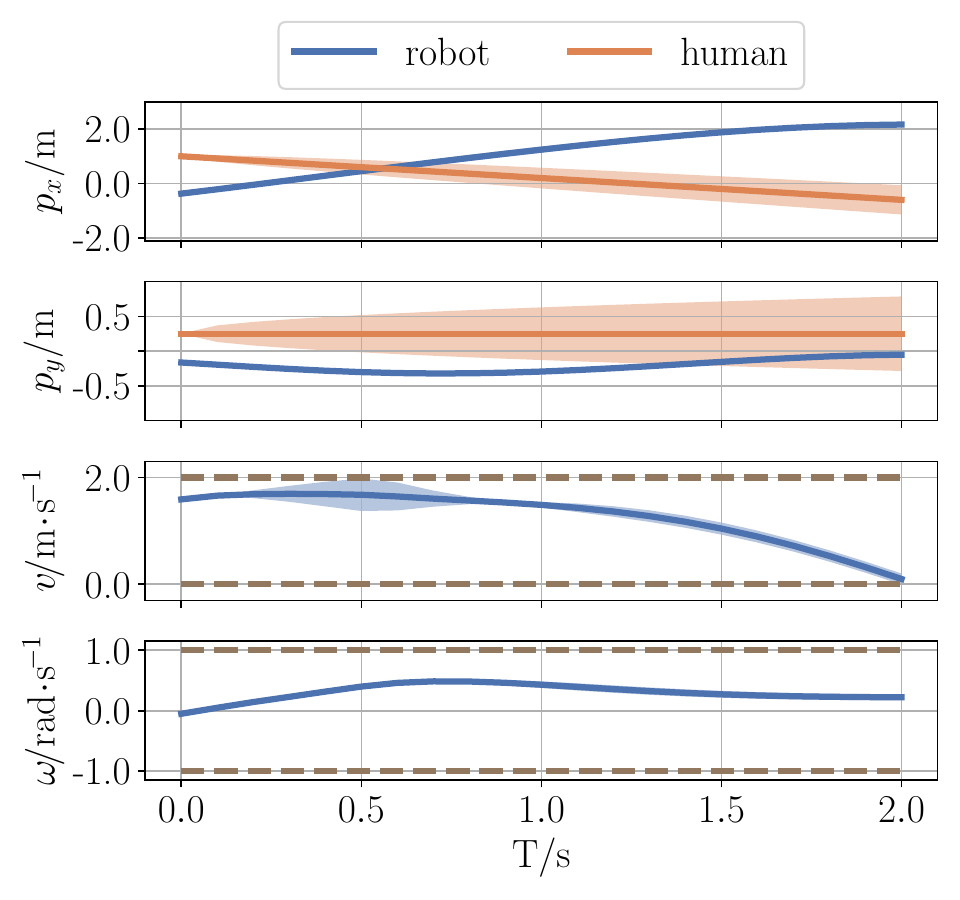}};
	    	\node at(a.center)[draw, red, dotted, line width=1.5pt,ellipse, minimum width=40pt, minimum height=15pt,rotate=0,yshift=-6pt,xshift=-20pt]{};
	    \end{tikzpicture}
	    \label{fig:ocp_solution_OPT_corridor_time_whole_044}}
    \vspace{-5pt}
	\caption{Planned robot state trajectories in time.
		Full-state feedback is incorporated.
		The shadows represent uncertainty tubes (three standard deviations).
		The red ellipsoid is for highlighting purpose.
	}
	 \label{fig:ocp_traj_in_time}
	\end{minipage}
    \hfill
	\begin{minipage}[b]{0.38\linewidth}
		\centering
		\includegraphics[width=\textwidth]{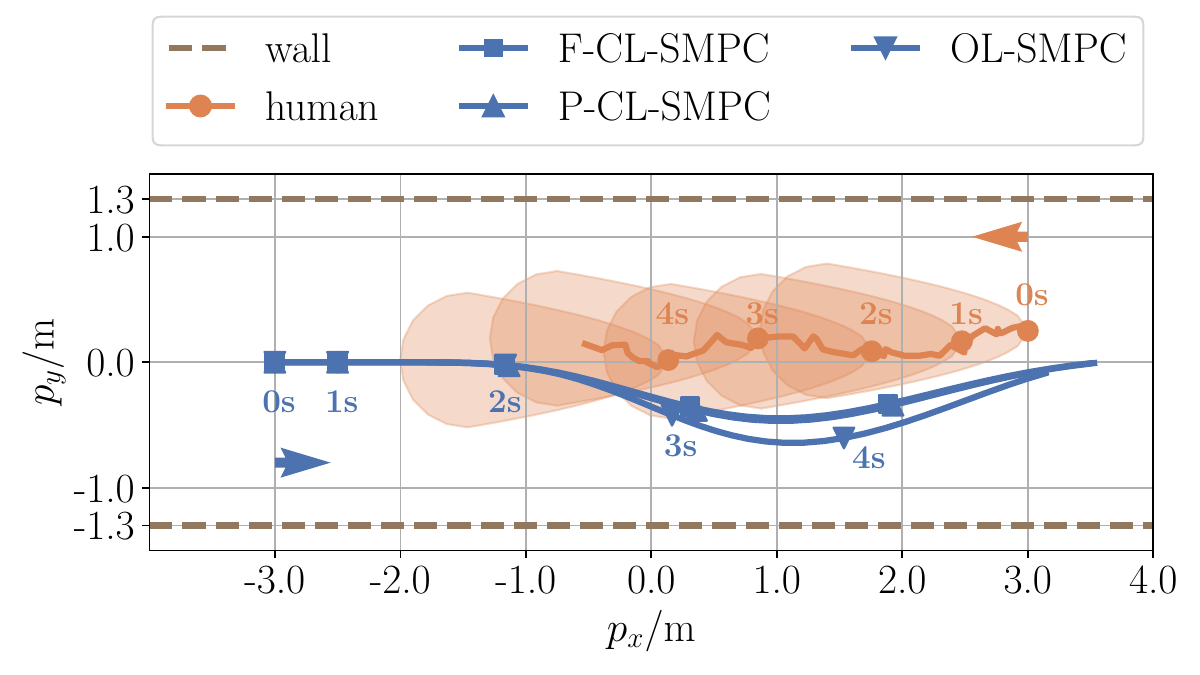}
		\vspace{-20pt}
		\caption{Robot and human trajectories in an MPC simulation of five seconds.
			The blue markers depict the robot positions at certain time instants.
			The orange semi-transparent sets correspond to the modeled human occupancy in space (three standard deviations).}
		\label{fig:position-mpc-simulation-disturbed}
	\end{minipage}
\end{figure*}

We first present the OCP planned trajectories.
We subsequently evaluate the MPC trajectories simulated based on different realizations of human motion uncertainty.
The assessment of computation time is presented at the end.
We implement the OCPs in CasADi \citep{Andersson2019}. We use the solver IPOPT~\citep{Waechter2006} to solve the nonlinear programming problems with MA-27~\citep{HSL} as the linear solver.
The computations are carried out on a computer with an Intel(R) Xeon(R) W-1250P CPU and 32GB of RAM.

\vspace{-5pt}
\subsection{OCP Planned Trajectories}
\vspace{-8pt}
\label{sec:testing-ocp}
\subsubsection*{Test Scenario}
We simulate a robot driving in a corridor.
The robot reference trajectory is traveling along the $x$-axis at a constant velocity:
\begin{equation*}
	x^{\mathrm{r,ref}}_{k} \! = \! {\left[  \begin{array}{ccccc}
		\left( \bar{\bar{p}}^{\mathrm{r}}_x + v_x^{\mathrm{r,ref}}\cdot ( t^{\mathrm{sim}}\! + \! k \discInterval )\right)  & 0 & 0 & v_x^{\mathrm{r,ref}} & 0
	\end{array} \right] }^\top,
\end{equation*}
for $k \!\in\! \mathbb{N}_{\left[0, N\right]}$, where $\bar{\bar{p}}^{\mathrm{r}}_x$ is the robot $x$-axis position at the beginning of the simulation,
   $t^{\mathrm{sim}}$ is the time in simulation,
   and $v_x^{\mathrm{r,ref}}$ is the reference velocity.
A human walks in the corridor in the opposite direction and the nominal velocity is constant:
$x^{\mathrm{h}}_{k} \!= \! {\left[  \begin{array}{cc}
			\left( \bar{\bar{p}}^{\mathrm{h}}_x \!+\! {v}^{\mathrm{h}}_x \cdot ( t^{\mathrm{sim}}\! + \! k \discInterval ) \right)
			 &  \bar{\bar{p}}^{\mathrm{h}}_y
		\end{array} \right] } ^{\!\top}$, for $k \!\in\! \mathbb{N}_{\left[0, N\right]}$,
where $\left[  \begin{array}{cc}
	 \bar{\bar{p}}^{\mathrm{h}}_x   &  \bar{\bar{p}}^{\mathrm{h}}_y
\end{array} \right]^{\!\top}$
is the human state at the start of the simulation.
\yunfan{Although the nominal human velocity ${u}^{\mathrm{h}}_k$ is constant in this test scenario, the velocity ${u}^{\mathrm{h}}_k$ can be time-varying.
The determination of the values of ${u}^{\mathrm{h}}_k$ alongside the uncertainty matrix $W_k^{\mathrm{h}}$ is not the focus of this paper.}
The robot state and the human state are exactly known at the initial state of the OCP.
The OCP parameters are collected in Table~\ref{table:param-robot-example}.

\begin{table}[t]
	\centering
	\captionsetup{width=\linewidth}
	\caption{Parameters used in simulation in Sec.~\ref{sec:results}}
	\label{table:param-robot-example}
	\vspace{-4pt}
	\begin{tabular}{@{} l  r  r r @{}}
		\hline
		\hline
		& & & \\[\dimexpr-\normalbaselineskip+2pt]
		Name              & Unit         & Symbol     & Value  \\
		\hline
		& & & \\[\dimexpr-\normalbaselineskip+2pt]
		Discretization intervals  &     & $N$     & 20    \\
		Safety distance&m             & $\safeR$ & 0.3 \\
		Human velocity covariance & m${}^{2} \cdot$s${}^{\shortminus 2}$   & $W_k^{\mathrm{h}}$   & $0.4^2\cdot I_2$ \\
		Weight of position tracking error & m${}^{\shortminus 2}$          &    & 50  \\
		Weight of velocity tracking error & s${}^{2} \cdot$m${}^{\shortminus 2}$       &    & 2   \\
		Weight of input tracking error & s${}^{4} \cdot$m${}^{\shortminus 2}$   &    & 2   \\
		\hline
		\hline
	\end{tabular}
\end{table}

Examples of the OCP planned trajectories for the number of standard deviations $\gamma$ equal to three are shown in Fig.~\ref{fig:ocp_traj_in_space} and Fig.~\ref{fig:ocp_traj_in_time}.
Note that the overlap between the orange human ellipsoids and the blue semi-transparent sets
does \textit{not} imply that the collision chance constraints are violated.
The overlap is due to the correlation between the robot state uncertainty and human state uncertainty arising from the feedback scheme.
The correlation indicates that the robot will respond to the human motion uncertainty.

When the robot meets the human at the end of the prediction horizon,
    the optimized feedback policy counteracts human motion uncertainty by adapting the robot angular acceleration,
    particularly the uncertainty in the direction perpendicular to the robot forward motion.
It enables the robot to stay closer to the reference trajectory  (see Fig.~\ref{fig:ocp_solution_OPT_corridor_space_whole_188}).
Meanwhile, the robot gains uncertainty, particularly in its angular velocity (see Fig.~\ref{fig:ocp_solution_OPT_corridor_time_whole_188}).

In the scenario that the robot manages to pass the human in the prediction horizon,
    the robot applies feedback while passing by the human,
    especially via the forward acceleration.
It results in uncertainty in the robot forward velocity,
    as highlighted by the dotted-line red ellipsoid in Fig.~\ref{fig:ocp_solution_OPT_corridor_time_whole_044}.
After the robot safely passes the human,
the full-state feedback enables the robot to reduce its own state uncertainty through feedback.
In consequence, the robot state turns out relatively certain at the end of the prediction horizon (see Fig.~\ref{fig:ocp_solution_OPT_corridor_space_whole_044}).
In contrast, in the case of partial-state feedback,
    the robot state, especially the robot position, is remarkably more uncertain (see Fig.~\ref{fig:ocp_solution_OPT_corridor_space_partial_042}).

\vspace{-5pt}
\subsection{MPC Simulation}
\vspace{-5pt}
The human and robot trajectories running open-loop and closed-loop stochastic MPCs are shown in Fig~\ref{fig:position-mpc-simulation-disturbed}.
The robot steered by open-loop stochastic MPC (OL-SMPC) deviates more from the reference trajectory and is slower compared to closed-loop stochastic MPC (CL-SMPC).
The difference of counteracting solely the human state uncertainty (P-CL-SMPC) versus both the human and the robot uncertainty (F-CL-SMPC) is comparatively small.

We test the collision likelihood and cost values given 1000 human trajectory samples
    for $\numStd=3.0$ and $\numStd=2.0$ respectively.
The numerical solver occasionally fails to converge to a solution.
In such cases, we use the control policies optimized in previous time steps.
In each five-second long simulation,
the robot and human meet once.
One simulation is aborted once a collision takes place.
For each simulation, we compute the mean of the trajectory-tracking stage cost~\eqref{eq:tracking-stage-def} evaluated for $k=0$ at simulated time steps.
The mean stage cost and the minimum robot-human distance of different simulations are shown in Fig.~\ref{fig:minDist_cost}.
The median of the mean stage cost and the number of collisions over 1000 simulations are reported in Table~\ref{tab:mpc-collision}.

Nominal MPC neglects human motion uncertainty and therefore shows a high likelihood of collision.
OL-SMPC attains low collision frequency
    while keeping an overly large distance to the human (see the markers in the dotted-line red ellipsoids in Fig.~\ref{fig:minDist_cost}).
This results in, on average, a relatively high trajectory-tracking cost (see Table.~\ref{tab:mpc-collision}).
The stage cost goes high
    when the robot gets stuck.
Such cases are more frequent when the robot is steered by OL-SMPC compared with CL-SMPC (see Fig.~\ref{fig:minDist_cost}).
The tracking performance associated with P-CL-SMPC is similar to that of F-CL-SMPC, as measured by the trajectory-tracking stage cost.

\begin{figure*}[t]
	\centering
	\begin{minipage}{0.6\linewidth}
		\centering
		\captionsetup[subfloat]{captionskip=-2pt}
		{\begin{tikzpicture}
			\node(a){\includegraphics[width=\linewidth]{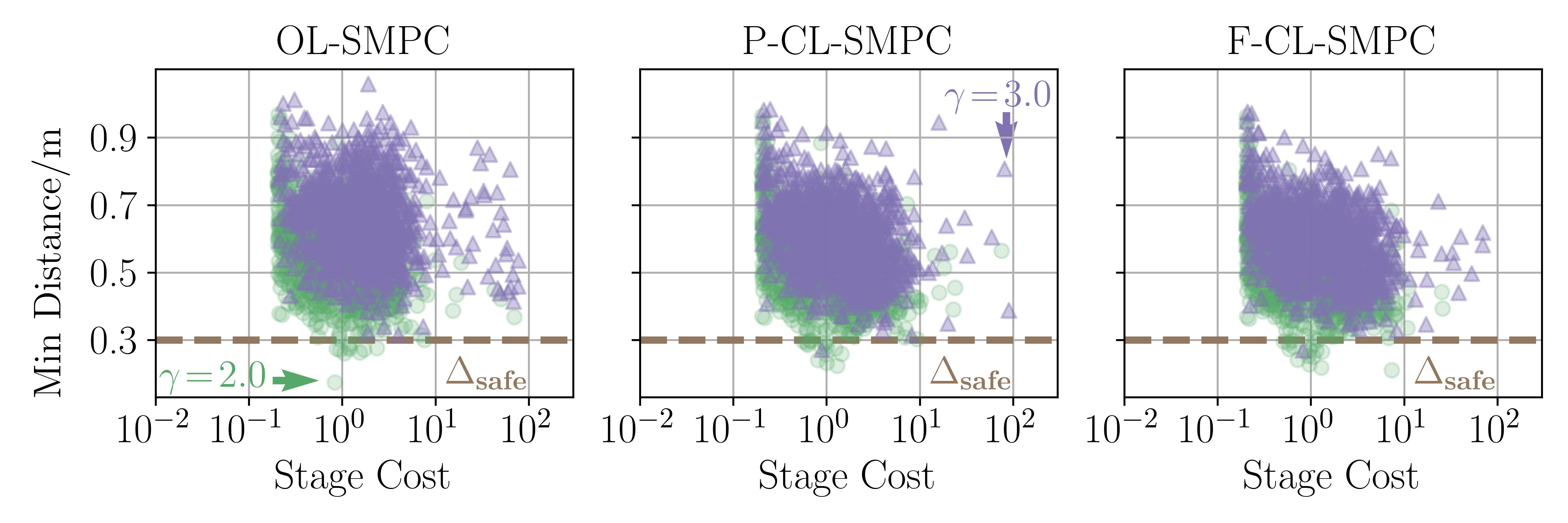}
			};
			\node at(a.center)[draw, red, dotted, line width=2pt, ellipse, minimum width=35pt, minimum height=18, rotate=0, xshift=-80pt, yshift=23pt]{};
		\end{tikzpicture}
		\vspace{-30pt}
		\caption{Mean stage cost and minimum robot-human distance.
		Purple triangles and green circles correspond to $\numStd=3.0$ and $\numStd=2.0$.}
		\label{fig:minDist_cost}
		}
	\end{minipage}
	\hfill
	\begin{minipage}{0.37\linewidth}
		\centering
		\captionsetup[subfloat]{captionskip=-2pt}
		\includegraphics[width=\linewidth]{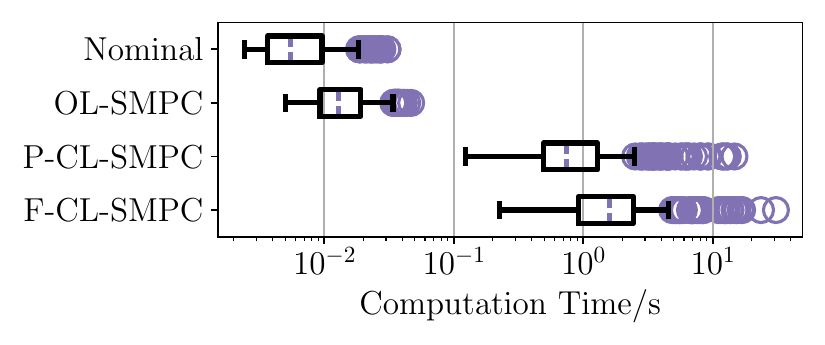}
		\vspace{-22pt}
		\caption{Computation time of solving OCPs ($\numStd=3.0$).
		The box extends from the lower to upper quartile of the data. The dashed line shows the median.}
		\label{fig:wall_time_histgram_partial_fdbk}
	\end{minipage}
	\vspace{-3pt}
\end{figure*}

\subsection{Computation Time Evaluation}
\vspace{-9pt}
In addition to the straight-line corridor scenario,
    the robot is anticipated to navigate curved paths.
The curvature of the reference trajectory potentially impacts the degree of nonlinearity involved in the stochastic OCP.
Therefore, for the evaluation of computation time, we employ a modified test scenario:
    the robot navigates along a circular arc while
the human participant travels along a concentric arc.
Our experiment involves solving 300 OCPs,
    with variations in the starting and the goal positions of the robot as well as the curvature of the circular arcs.

\begin{table}[b]
	    \captionsetup{width=\linewidth}
		\centering
		\caption{Number of occurred collisions and median of the mean stage cost over 1000 simulations.}
		\label{tab:mpc-collision}
	\begin{tabular}{l l@{\hskip 20pt}r@{\hskip 20pt}r@{}}
		\hline \hline
		& & &\\[\dimexpr-\normalbaselineskip+2pt]
		&     & \# collisions & Stage cost\,\eqref{eq:tracking-stage-def} \\
		\hline \hline
		\multicolumn{1}{l|}{}	& & &\\[\dimexpr-\normalbaselineskip+2pt]
		\multicolumn{1}{l|}{}                     & Nominal MPC & 305                                                     &                                      0.25    \\
		\hline
		\multicolumn{1}{l|}{}	& & &\\[\dimexpr-\normalbaselineskip+2pt]
		\multicolumn{1}{@{}l|}{\multirow{3}{*}{$\numStd=3.0$}} & OL-SMPC          & 0                                                       & 1.66                                               \\
		\multicolumn{1}{l|}{}         & P-CL-SMPC     & 1           & 1.29                \\
		\multicolumn{1}{l|}{}         & F-CL-SMPC     & 1           & 1.27                \\ \hline
		\multicolumn{1}{l|}{}	& & &\\[\dimexpr-\normalbaselineskip+2pt]
		\multicolumn{1}{@{}l|}{\multirow{3}{*}{$\numStd=2.0$}} & OL-SMPC     & 10                                                       & 0.98                                               \\
		\multicolumn{1}{l|}{}           &  P-CL-SMPC     & 17         & 0.88                                               \\
		\multicolumn{1}{l|}{}           &  F-CL-SMPC     & 18         & 0.84                                               \\
		\hline
		\hline
	\end{tabular}
\end{table}

The initial guess of the state variables for the nominal OCP is derived from the reference trajectory,
    for all other OCPs the initial guess is derived from the solution of the nominal OCP.
The computation time is reported in Fig.~\ref{fig:wall_time_histgram_partial_fdbk}.
\yunfan{For OL-SMPC,
    the covariance propagation can be precomputed.
The covariance matrices are passed to the OCPs as optimization variables
	while the constraint uncertainties $\beta_k$ of nonlinear constraints are optimized within the OCPs.}
The median computation time for nominal OCPs, OL-SMPC, P-CL-SMPC, and F-CL-SMPC is 5.5 milliseconds, 13 milliseconds, 0.75 seconds, and 1.6 seconds respectively.

\rem{
There still exist cases that the numerical solver fails to converge although the OCP is feasible.
Further improvement of numerical performance is subject to future research.
One practical approach is to adopt a homotopy optimization scheme,
    where we start with solving a nominal OCP
    and gradually increase the scale of the uncertainty until we reach the scale of the modeled uncertainty.
This approach improves the reliability of the solver at the cost of increasing the computation time.
Although the computation time is relatively long,
    it is possible to use the optimized feedback policy for a few time steps while the new OCP is being solved.
}

\vspace{-5pt}
\section{Conclusion}
\vspace{-8pt}
This paper presents a stochastic-MPC-based method for robot motion planning which incorporates feedback on human motion uncertainty.
The OCP planned trajectories and MPC trajectories of different feedback policies are compared in simulation.
The partial feedback policy overall strikes a favorable balance between computational speed and performance.
Future work could focus on modeling the human feedback on the robot motion
and efficient algorithms to solve the closed-loop stochastic OCP.


\vspace{-3pt}
\bibliography{refs} 

\begin{thebibliography}{20}
\providecommand{\natexlab}[1]{#1}
\providecommand{\url}[1]{\texttt{#1}}
\providecommand{\urlprefix}{URL }
\expandafter\ifx\csname urlstyle\endcsname\relax
  \providecommand{\doi}[1]{doi:\discretionary{}{}{}#1}\else
  \providecommand{\doi}{doi:\discretionary{}{}{}\begingroup
  \urlstyle{rm}\Url}\fi

\bibitem[{Andersson et~al.(2019)Andersson, Gillis, Horn, Rawlings, and
  Diehl}]{Andersson2019}
Andersson, J.A.E., Gillis, J., Horn, G., Rawlings, J.B., and Diehl, M. (2019).
\newblock {CasADi} -- a software framework for nonlinear optimization and
  optimal control.
\newblock \emph{Math. Program. Comput.}, 11(1), 1--36.

\bibitem[{Brock and Khatib(2002)}]{Brock2002}
Brock, O. and Khatib, O. (2002).
\newblock Elastic strips: A framework for motion generation in human
  environments.
\newblock \emph{Int. J. Robotics Res.}, 21(12), 1031--1052.

\bibitem[{Carvalho et~al.(2014)Carvalho, Gao, Lefevre, and
  Borrelli}]{Carvalho2014}
Carvalho, A., Gao, Y., Lefevre, S., and Borrelli, F. (2014).
\newblock {S}tochastic predictive control of autonomous vehicles in uncertain
  environments.
\newblock In \emph{Proc. 12th Int. Symp. Adv. Vehicle Control}.

\bibitem[{Castillo-Lopez et~al.(2020)Castillo-Lopez, Ludivig, Sajadi-Alamdari,
  Sanchez-Lopez, Olivares-Mendez, and Voos}]{Castillo-Lopez2020}
Castillo-Lopez, M., Ludivig, P., Sajadi-Alamdari, S.A., Sanchez-Lopez, J.L.,
  Olivares-Mendez, M.A., and Voos, H. (2020).
\newblock A real-time approach for chance-constrained motion planning with
  dynamic obstacles.
\newblock \emph{IEEE Robotics and Automat. Lett.}, 5(2), 3620--3625.

\bibitem[{Du~Toit and Burdick(2012)}]{DuToit2012}
Du~Toit, N.E. and Burdick, J.W. (2012).
\newblock {R}obot motion planning in dynamic, uncertain environments.
\newblock \emph{{IEEE} Trans. Robotics}, 28(1), 101--115.

\bibitem[{Goulart et~al.(2006)Goulart, Kerrigan, and Maciejowski}]{Goulart2006}
Goulart, P.J., Kerrigan, E.C., and Maciejowski, J.M. (2006).
\newblock Optimization over state feedback policies for robust control with
  constraints.
\newblock \emph{Automatica}, 42, 523--533.

\bibitem[{Han and Mangasarian(1979)}]{Han1979}
Han, S.P. and Mangasarian, O.L. (1979).
\newblock Exact penalty functions in nonlinear programming.
\newblock \emph{Math. Program.}, 17(1), 251–269.

\bibitem[{{HSL}(2011)}]{HSL}
{HSL} (2011).
\newblock {A} collection of {F}ortran codes for large scale scientific
  computation.
\newblock \url{http://www.hsl.rl.ac.uk}.

\bibitem[{Macek et~al.(2008)Macek, Vasquez, Fraichard, and
  Siegwart}]{Macek2008}
Macek, K., Vasquez, D., Fraichard, T., and Siegwart, R. (2008).
\newblock {T}owards safe vehicle navigation in dynamic urban scenarios.
\newblock \emph{Automatika}, 50, 184--194.

\bibitem[{Mayne et~al.(2011)Mayne, Kerrigan, and Falugi}]{Mayne2011a}
Mayne, D.Q., Kerrigan, E.C., and Falugi, P. (2011).
\newblock Robust model predictive control: advantages and disadvantages of
  tube-based methods.
\newblock \emph{Proc. IFAC World Congr.}, 44(1), 191--196.

\bibitem[{Messerer et~al.(2023)Messerer, Baumg{\"a}rtner, and
  Diehl}]{Messerer2023}
Messerer, F., Baumg{\"a}rtner, K., and Diehl, M. (2023).
\newblock A dual-control effect preserving formulation for nonlinear
  output-feedback stochastic model predictive control with constraints.
\newblock \emph{IEEE Contr. Syst. Lett.}, 7, 1171--1176.

\bibitem[{Messerer and Diehl(2021)}]{Messerer2021}
Messerer, F. and Diehl, M. (2021).
\newblock An efficient algorithm for tube-based robust nonlinear optimal
  control with optimal linear feedback.
\newblock In \emph{Proc. IEEE Conf. Decis. Control (CDC)}.

\bibitem[{Nagy and Braatz(2004)}]{Nagy2004}
Nagy, Z. and Braatz, R. (2004).
\newblock {O}pen-loop and closed-loop robust optimal control of batch processes
  using distributional and worst-case analysis.
\newblock \emph{J. Process Control}, 14, 411--422.

\bibitem[{Schwarm and Nikolaou(1999)}]{Schwarm1999}
Schwarm, A. and Nikolaou, M. (1999).
\newblock Chance‐constrained model predictive control.
\newblock \emph{AIChE J.}, 45(8), 1743--1752.

\bibitem[{Schwarting et~al.(2018)Schwarting, Alonso-Mora, Paull, Karaman, and
  Rus}]{Schwarting2018}
Schwarting, W., Alonso-Mora, J., Paull, L., Karaman, S., and Rus, D. (2018).
\newblock Safe nonlinear trajectory generation for parallel autonomy with a
  dynamic vehicle model.
\newblock \emph{IEEE Trans. Intell. Transp. Syst.}, 19(9), 2994--3008.

\bibitem[{Scokaert and Mayne(1998)}]{Scokaert1998}
Scokaert, P.O.M. and Mayne, D.Q. (1998).
\newblock Min-max feedback model predictive control for constrained linear
  systems.
\newblock \emph{{IEEE Trans.\ Automat.\ Contr.}}, 43, 1136--1142.

\bibitem[{Siegwart et~al.(2011)Siegwart, Nourbakhsh, and
  Scaramuzza}]{Siegwart2011}
Siegwart, R., Nourbakhsh, I.R., and Scaramuzza, D. (2011).
\newblock \emph{Introduction to autonomous mobile robots}.
\newblock MIT Press, Cambridge, MA, USA.

\bibitem[{W\"achter and Biegler(2006)}]{Waechter2006}
W\"achter, A. and Biegler, L.T. (2006).
\newblock On the implementation of an interior-point filter line-search
  algorithm for large-scale nonlinear programming.
\newblock \emph{Math. Program.}, 106(1), 25--57.

\bibitem[{Yan and Bitmead(2005)}]{Yan2005}
Yan, J. and Bitmead, R. (2005).
\newblock {I}ncorporating state estimation into model predictive control and
  its application to networktraffic control.
\newblock \emph{Automatica}, 41, 595--604.

\bibitem[{Zhu and Alonso-Mora(2019)}]{Zhu2019}
Zhu, H. and Alonso-Mora, J. (2019).
\newblock {C}hance-constrained collision avoidance for {MAV}s in dynamic
  environments.
\newblock \emph{IEEE Robotics and Automat. Lett.}, 4(2), 776--783.

\end{thebibliography}

\end{document}